\documentclass[manuscript]{acmart} 
\renewcommand\footnotetextcopyrightpermission[1]{}
\AtBeginDocument{%
  \providecommand\BibTeX{{%
    \normalfont B\kern-0.5em{\scshape i\kern-0.25em b}\kern-0.8em\TeX}}}

\usepackage{amsmath,amsfonts,mathrsfs,mathtools}
\usepackage{algorithm, algpseudocode}
\usepackage{subfig}
\usepackage{xcolor}
\usepackage{booktabs}
\usepackage{multirow}

\begin{document}

\title{Momentum-based Gradient Methods in Multi-Objective Recommendation}
\titlenote{Copyright 2021 for this paper by its authors. Use permitted under Creative Commons License Attribution 4.0 International (CC BY 4.0).\\Presented at the MORS workshop held in conjunction with the 15th ACM Conference on Recommender Systems (RecSys), 2021, in Amsterdam, Netherlands.}
\author{Blagoj Mitrevski}
\authornote{Work done while at EPFL and Swisscom.}
\affiliation{%
  \institution{Symphony}
  \country{North Macedonia}
  }
\email{blagoj.mitrevski@symphony.is}

\author{Milena Filipovic}
\authornotemark[1]
\affiliation{%
  \institution{Swisscom}
  \country{Switzerland}
  }
\email{milena.filipovic@swisscom.com}

\author{Diego Antognini}
\affiliation{%
 \institution{Ecole Polytechnique Fédérale de Lausanne}
 \country{Switzerland}
 }
\email{diego.antognini@epfl.com}

\author{Emma Lejal Glaude}
\affiliation{%
  \institution{Swisscom}
  \country{Switzerland}
  }
\email{emma.lejal.glaude@swisscom.com}

\author{Boi Faltings}
\affiliation{%
 \institution{Ecole Polytechnique Fédérale de Lausanne}
 \country{Switzerland}
 }
\email{boi.faltings@epfl.com}

\author{Claudiu Musat}
\affiliation{%
  \institution{Swisscom}
  \country{Switzerland}
  }
\email{claudiu.musat@swisscom.com}

\renewcommand{\shortauthors}{Mitrevski, et al.}

\begin{abstract}
Multi-objective gradient methods are becoming the standard for solving multi-objective problems. Among others, they show promising results in developing multi-objective recommender systems with both correlated and conflicting objectives. Classic multi-gradient~descent usually relies on the combination of the gradients, not including the computation of first and second moments of the gradients. This leads to a brittle behavior and misses important areas in the solution space. 
In this work, we create a multi-objective model-agnostic \textit{Adamize} method that leverage the benefits of the Adam optimizer in single-objective problems.
This corrects and stabilizes~the~gradients of every objective before calculating a common gradient descent vector that optimizes all the objectives simultaneously. We evaluate the benefits of \textit{Multi-objective Adamize} on two multi-objective recommender systems and for three different objective combinations, both correlated or conflicting. We report significant improvements, measured with three different Pareto front metrics: hypervolume, coverage, and spacing. Finally, we show that the \textit{Adamized} Pareto front strictly dominates the previous one on multiple objective pairs.
\end{abstract}




\maketitle

\section{Introduction}
\textit{Decision-making relies on multiple factors.} The world is complex, and many problems require an optimization for more than one objective. Multi-objective problems are present in fields like engineering, economics, finance, logistics, and many more. Multi-objective optimization is the area of decision-making in which we simultaneously optimize for more than one objective. We distinguish two types of objectives: the correlated and the conflicting ones. When the objectives are conflicting, the choice of the optimal decisions needs to be taken in the presence of trade-offs: choosing one objective usually comes at the expense of the others. In practice, the decision of choosing the best solution is left to the domain experts or the business stakeholders. Multi-objective optimization provides a data-driven alternative.

\textit{Recommenders are not only about relevance.}
One of the objectives of recommender systems is to be accurate, namely to successfully model the user's preferences. These systems are however not limited to accuracy. Another objective that can improve the user's experience with the recommender system is proposing more diverse content. It helps the user to escape their filter bubble that can reduce user creativity, learning, and connection \cite{nguyen2014exploring}. Also, promoting more recent content \cite{chakraborty2017optimizing, gabriel2019contextual} can bring social value by keeping the user up to date.

However, among the multiple stakeholders of the recommender system it is possible to encounter diverse and competing objectives. For instance, increasing the revenue for a company does not always mean the user will get a better and improved experience. If an application store puts more importance on recommending overpriced applications it may increase its revenue, but this strategy will hurt developers of free and cost-effective applications; also it will put a burden on the user's budget. This becomes more frequent, as more and more companies are becoming socially responsible \cite{varona2020incentives,hatcher2000social,vveinhardt2014readiness}, which can be unaligned with traditional business objectives. 

\textit{From one to multiple objectives.} 
Prior work \cite{poirion2017descent} proposed the gradient-based multi-objective optimization algorithm, called the Stochastic Multi-Subgradient Descent Algorithm (SMSGDA), or an improved version for recommendation~\cite{milojkovic2019multi}. The method computes the gradients of each objective and then constructs a common descent vector by taking a linear combination of the individual gradients. Each gradient's weight is computed by solving a quadratic constrained~optimization problem. Finally, the model parameters are updated in the opposite direction of the common descent vector. The problem with stochastic optimization is the stochasticity that comes from using mini-batches or dropout regularization. 

In single-objective settings, this problem is solved using optimizers like Adam \cite{kingma2014adam} and RMSprop \cite{hinton2012neural}. These stabilize the computation and speed up to convergence. In a similar fashion, we introduce a simple yet effective \textit{Adamize} trick for multi-objective problems.  We keep track of the first and second moments of the gradients and use the momentums to correct the gradients and compute better gradient weights. Finally, we calculate a more stable common descent vector using the corrected gradients. 

\textit{In this work, we thus make the following contributions:} we address the recommendation task with multiple-objectives, in which objectives can either be correlated or conflicting. We first present the \textit{Adamize} trick to correct and stabilize the gradients of every objective before aggregating them into a common gradient descent. 
We then show that our novel multi-gradient descent method is model-agnostic and can be easily integrated into state-of-the-art recommender systems. We evaluate our method using two real-world recommendation datasets with up to three objectives. We then compare the results of the momentum-based optimization with the state-of-the-art using three different metrics based~on the resulting Pareto fronts. As the observed differences are stark, we complement our analysis with visualizations that further underline the usefulness of momentum-based multi-gradient descent in multi-objective recommender~systems.

\section{Related work}

With the advances of neural approaches in other fields, they also found their way into recommendation systems. First, the introduction of Neural network-based Collaborative Filtering \cite{he2017neural} showed promising results. Later, the Variational Autoencoders for Collaborative Filtering \cite{liang2018variational} became state-of-the-art and still keeps its title as one of the best collaborative filtering based recommender.

Recommender systems and ranking problems have similarities: learning a personalized recommender can be transformed as a ranking problem \cite{karatzoglou2013learning}. The multi-objective ranking optimization in \cite{carmel2020multi} is solved by label aggregation. This method collects the multiple labels of the training examples into a single label, and then use a single-objective optimizer to rank the aggregated label, solving the multi-objective problem.

Alternatively, the gradient-based methods can solve the multi-objective optimization problem. In \cite{desideri2012multiple}, the authors propose the Multi-Gradient Descent Algorithm (MGDA) for optimizing multi-objective based on the steepest descent method. This algorithm is an adjustment of the classical gradient descent algorithm to work with multiple objectives. The same authors of the MGDA algorithm extended it to Stochastic Multi-Subgradient Descent Algorithm (SMSGDA) \cite{poirion2017descent}. The SMSGDA is a stochastic version of the MGDA that could also work with non-smooth objective functions. A more robust gradient-based multi-objective optimization algorithm that still works in cases when the exact gradients could not be computed is presented in \cite{peitz2018gradient}. To alleviate the inaccuracies, an additional condition is presented for the descent direction.
\cite{milojkovic2019multi} proposed a gradient-based algorithm for optimizing multi-objective recommender systems. Their solution is based on finding a common descent vector, which is a combination of the gradients of every objective. By taking an optimization step in the opposite direction of this common descent vector, the model is optimized for all objectives simultaneously. We build upon this work and improve convergence and stability of the optimization.

\section{Background}
There are different ways of solving the multi-objective optimization problem, such as re-ranking or gradient-based solutions. In this work, we focus on the latter and base our work on the multi-gradient descent algorithm \cite{milojkovic2019multi}.

\subsection{Definitions}

\subsubsection{Multi-Objective Optimization.}
The multi-objective optimization of a model can be formally defined as:\begin{equation}
\min_{w \in \mathbb{R}^D}{\mathcal{L}(w)} = \min_{w \in \mathbb{R}^D}
\mathcal{L}_1(w), \dots
\mathcal{L}_n(w)  
\end{equation}where $w$ are the model parameters, $D$ is the dimension of the model parameters, $\mathcal{L}(w): \mathbb{R}^D \rightarrow \mathbb{R}^n$ is a vector valued objective function with continuously differentiable objective
functions $\mathcal{L}_n(w): \mathbb{R}^D \rightarrow \mathbb{R}$.

\subsubsection{Common Descent Vector.}
The common descent vector \cite{desideri2012multiple} is the core of the multi-gradient descent algorithm. It is computed with a linear combination of the gradients:
\begin{equation}
\label{eq:cdv}
    \nabla_{w} \mathcal{L}(w)=\sum_{i=1}^{n} \alpha_{i} \nabla_{w} \mathcal{L}_{i}(w)
\end{equation}
with $\alpha_{i} \geq 0, i \in \{1,\dots,n\}$, and $\sum_{i=0}^{n}{\alpha_{i}} = 1$, where $\mathcal{L}_{i}(w)$ is the gradient of the i-th objective, $\alpha_{i}$ is the weight of the i-th gradient objective, $n$ is the number of objectives, and $w$ are the model parameters.


\subsubsection{Pareto Stationary Solution.} A solution $w$ of the eq. (\ref{eq:cdv}) is Pareto stationary iff it satisfies the Karush–Kuhn–Tucker (KKT) conditions. In other words, there exists $\alpha_{1} \dots \alpha_{n}$ that statisfy the three following constraints:
$$
\alpha_{1} \dots \alpha_{n} \geq 0
\text{,} 
\sum_{i=0}^{n}{\alpha_{i}} = 1
\text{, and}
\sum_{i=1}^{n} \alpha_{i} \nabla_{w} \mathcal{L}_{i}(w)=0
$$

\subsection{Multi-Gradient Descent Algorithm (MGDA)}
\label{mgda_subsection}
After the definition of the common descent vector and the Pareto stationary solution, we present the multi-gradient descent algorithm (MGDA) \cite{desideri2012multiple}. The algorithm is deterministic and is proven to converge to a Pareto stationary solution. For an arbitrary number of objectives, this algorithm computes the alphas (i.e., weights of gradients, see eq. (\ref{eq:cdv})) to create a common descent vector. This vector is made such that the optimization step in the opposite direction of this common descent vector; all the objectives are simultaneously optimized. To compute the alphas, we need to solve the following quadratic constrained optimization problem~(QCOP):
\begin{equation}
    \min _{\alpha_{1}, \ldots, \alpha_{n}}\left\{\left\|\sum_{i=1}^{n} \alpha_{i} \nabla_{w} \mathcal{L}_{i}(w)\right\|^{2} | \sum_{i=1}^{n} \alpha_{i}=1, \alpha_{i} \geq 0\right\}
\end{equation}
After computing the alphas, we compute the final common descent vector $\nabla_{w} \mathcal{L}(w)$. If $\nabla_{w} \mathcal{L}(w) = 0$ the solution is Pareto Stationary. Otherwise, $\nabla_{w} \mathcal{L}(w) \neq 0$, the solution is not Pareto Stationary and thus, we apply an optimisation step in the opposite direction of the common descent vector, improving each objective at once.

It is worth noting that if there are two objectives, an analytical solution to the QCOP problem exists. Otherwise, the QCOP can be solved by using the Frank-Wolfe constrained optimization algorithm as in \cite{sener2018multi}.

\subsection{Stochastic Multi-Subgradient Descent Algorithm (SMSGDA)}
\label{SMSGDA}
The previous multi-gradient descent algorithm has few drawbacks to be used in real-world problems. A first one is the need to compute the full gradient at every optimization step which makes it computationally expensive and in some cases infeasible. A second one, the requirements do not allow to use non-smooth loss functions as objective functions. All of these drawbacks are solved by the Stochastic Multi-Subgradient Descent Algorithm (SMSGDA) presented in \cite{poirion2017descent}. The Stochastic Multi-Subgradient Descent Algorithm is similar to the Multi-Gradient Descent Algorithm, with the difference that instead of computing the gradients for every objective and then computing the alphas using the whole dataset, we are computing them on a subset of the dataset. Therefore, the stochasticity comes from using mini-batches.

\subsection{Gradient Normalization}
In real-world use-cases, the objectives for which we are optimizing may have different scales. This causes a problem for the MGDA and SMSGDA algorithms because they will favor the objectives that have a higher scale, leading to unbalanced solutions that perform well on certain objectives, but badly on the others. To solve this problem, after computing the gradients, the authors normalize them to interval according to the maximal empirical loss for each objective:
$
    \hat{\nabla_{w}\mathcal{L}_i(w)} = \frac{\nabla_{w} \mathcal{L}_i(w)}{\mathcal{L}_i(w_{init})}
$
,where $\hat{\nabla_{w}\mathcal{L}_i(w)}$ is the resulting normalized gradient, $\nabla_{w} \mathcal{L}_i(w)$ is the original gradient of the $i$-th objective, $\mathcal{L}_i(w_{init})$ is the initial loss for the $i$-th objective which is used as approximation for the maximum empirical loss for the given objective.


\section{The Adamize Trick for Multi-Objective Optimization}
\label{adamize_grads}
When optimizing models on a single objective, we are usually doing it in a stochastic fashion. The stochasticity comes from using mini-batch stochastic gradient descent where we use subsets of the data to compute the gradient, or use a dropout regularization \cite{srivastava2013improving}. The stochasticity in the optimization algorithm introduces noise in the gradient and may cause the algorithm to converge slower, or even diverge. 

There exist multiple optimizers like Adam \cite{kingma2014adam} and RMSprop \cite{hinton2012neural} which aim to stabilize the gradients when doing an optimization step. They achieve the stabilization by keeping a running average of the first and second moments of the gradients and taking a step in the opposite direction of the corrected gradient by using the first and the second momentum. For example, the corrected gradient moves faster on steep slopes and oscillates less on valleys and thus, move faster to the optima. Following the intuition behind ADAM and RMSprop, it may be beneficial, when using the Stochastic Multi-Subgradient Descent Algorithm (SMSGD) to smooth the gradients from the different objectives before calculating alphas and combining them to get the final common descent vector. Intuitively, this may lead to more stable alpha computations, faster convergence, and convergence to better solutions.\\
\indent The vanilla SMSGD algorithm is presented in Section~\ref{SMSGDA}. Our proposition is to use Adam based optimizers for every objective before computing the common descent vector. We directly add the Adam computation for every objective. Therefore, the difference with the vanilla SMSGD is that we are also keeping the running average for the gradient of every objective, instead of keeping only the average of the common descent vector. Since these are the gradients that affect the computations of the alphas, the final common descent vector is expected to be more stable. The pseudo-code of the \textit{Adamize} trick for the gradients is presented in Algorithm~\ref{Adamize1} and Algorithm~\ref{Adamize2}. The difference between the vanilla SMSGD and Algorithm~\ref{Adamize1} is in the bold line: instead of using the original gradients from every objective, we correct them using the first and second momentums, and we use the corrected gradients to compute the alphas and the common descent vector. To overcome the cold-start problem like in the original Adam algorithm, we initialize all the parameters to zero, and then update them very epoch.


In terms of computation and memory requirement, the complexity is linear with respect to the number of objectives. Memory-wise, we need a constant memory to save the first and second momentums of every objective. Furthermore,  the computation is constant with regard to the number of objectives, and adding an additional objective would require an additional call of the Adamize procedure, which does a constant computations with regards to the number of objectives. As the number of objectives is small, the overhead of our method is insignificant. Finally, this method is model-agnostic and can be used to optimize any model in a multi-objective fashion.

\begin{minipage}{0.46\textwidth}
\begin{algorithm}[H]
    \centering
    \scriptsize
    \caption{SMSGDA with Gradient Normalization and Adamizing Every Objective}
    \label{Adamize1}
    \begin{algorithmic}[1]
        \State $initialize()$
        \State $empirical\_loss_i = \mathcal{L}_i(w) \forall i \in {1,...,n}$
        \For{$epoch \in {1,...,M}$}
            \For{$batch \in {1,...,B}$}
                \State $forward\_pass()$; $evaluate\_model()$; $update\_pareto\_set()$
                \For{$i \in {1,...,n}$}
                    \State $calculate \, loss \:\;  \mathcal{L}_i(w)$
                    \State $calculate \, gradient \;\;  \nabla_{w} \mathcal{L}_i(w)$
                    \State $normalize \, gradient \;\;  \hat{\nabla_{w}\mathcal{L}_i(w)} = \frac{\nabla_{w} \mathcal{L}_i(w)}{empirical\_loss_i} $
                    \State $ \boldsymbol{\bar{\nabla_{w} \mathcal{L}_i(w)} = Adamize(\hat{\nabla_{w} \mathcal{L}_i(w)})}$
                \EndFor
                \State $\alpha_{1}, \ldots, \alpha_{n}=\operatorname{QCOPSolver}\left(\bar{\nabla_{w}\mathcal{L}_1(w)}, \ldots, \bar{\nabla_{w}\mathcal{L}_n(w)}\right)$
                \State $\bar{\nabla_{w} \mathcal{L}(w)}=\sum_{i=1}^{n} \alpha_{i} \bar{\nabla_{w}\mathcal{L}_i(w)}$
                \State $w = w - \eta \bar{\nabla_{w} \mathcal{L}(w)}$
            \EndFor
        \EndFor
    \end{algorithmic}
\end{algorithm}
\end{minipage}
\hfill
\begin{minipage}{0.46\textwidth}
\begin{algorithm}[H]
    \centering
    \scriptsize
    \caption{Adamizing a Gradient}
    \label{Adamize2}
    \begin{algorithmic}[1]
        \State \textbf{Parameters}: $\beta_1, \beta_2 \in [0, 1)$: Exponential decay rates for the moment estimates
        \State \textbf{Parameters}: $\lambda$: Gradient correction magnitude parameter
        \State $m_0 \leftarrow 0$ (Initialize 1st moment vector)
        \State $v_0 \leftarrow 0$ (Initialize 2nd moment vector)
        \State $t \leftarrow 0$ (Initialize timestep)
        
        \Procedure{Adamize}{$\nabla_{w} \mathcal{L}(w_t)$}
        \State $t \leftarrow t + 1$
        \State $g_t \leftarrow \nabla_{w} \mathcal{L}(w_t)$ (The gradient w.r.t. stochastic objective at timestamp t)
        \State $m_t \leftarrow \beta_1 * m_{t-1} + (1 - \beta_1) * g_t $ (Update biased first moment estimate)
        \State $v_t \leftarrow \beta_2 * v_{t-1} + (1 - \beta_2) * g^2_t $ (Update biased second raw moment estimate)
        \State $\hat{m_t} \leftarrow m_{t} / (1 - \beta^t_1) $ (Compute bias-corrected first moment estimate)
        \State $\hat{v_t} \leftarrow v_{t} / (1 - \beta^t_2) $ (Compute bias-corrected second raw moment estimate)
        \State \Return $(1-\lambda)\nabla_{w} \mathcal{L}(w_t) + \lambda * \hat{m_t} / (\sqrt{\hat{v_t}} + \epsilon) $ (Smoothed gradient)
        \EndProcedure
    \end{algorithmic}
\end{algorithm}
\end{minipage}

\section{Experiments}
\label{experimental_setup}
In this section, we assess the improvement of the proposed \textit{Adamize} trick. We follow the experimental design of \cite{milojkovic2019multi} and apply our method on recommender systems, although it can be used  to improve any multi-objective gradient based solution. We experiment on two datasets and up to three correlated and conflicting objectives.\footnote{For simplicity, we will use interchangeably the words objectives and losses.}

\subsection{Objectives for Recommendation}
\label{sec:obj_rec}
A recommender system can be trained with different objectives and for different purposes. For example, for some companies, there might be an economic or strategic incentive to recommend newer, instead of older, content to the users. Other socially responsible companies would like that their recommender to learn a notion of fairness or awareness of social biases. In this section, we present the objectives we employ in our experiments. As a use-case, we use the state-of-the-art variational autoencoder Mult-VAE\textsuperscript{PR} of \cite{liang2018variational} to demonstrate how to integrate our objectives into an existing recommender training procedure. However, we emphasize that they are easily adapted to any other model that, as recommendation, outputs a vector of probabilities across all the items.

\subsubsection{Relevance Objective.}
\label{relevance_obj}
This loss measures the relevance of the predicted items for the given user. The idea is to compare the output of the model with the user's interactions and measure how good the model can predict the user's interactions. The relevance loss in variational autoencoders is simply the reconstruction loss, plus the KL divergence between the posterior and the prior. More formally, the loss is:
\begin{equation}
\mathcal{L}(\boldsymbol{x};\theta,\phi) = \mathbb{E}_{q_\phi(\boldsymbol{z}|\boldsymbol{x})}[\log{p_\theta(\boldsymbol{x}|\boldsymbol{z})}] - \beta * KL(q_\phi(\boldsymbol{z}|\boldsymbol{x})||p(\boldsymbol{z}))
\end{equation}
where $\boldsymbol{x}$ is the input vector for a user, $\theta$ and $\phi$ are model parameters, $\textbf{z}$ is the variational parameter of the distribution, and $\beta$ is the regularizer controlling how much weight to be given to the KL term.

We use the definition of \cite{liang2018variational} to measure relevance. We quantify the ratio of relevant top-k items to users with~\textit{Recall@k}:
\begin{equation}
\label{eq:recallatk}
    Recall@k(u,\omega) := \frac{\sum_{r=1}^k\mathbb{I}[\omega(r) \in I_u]}{min(k, |I_u|)}
\end{equation}
where $\omega(r)$ is the item at rank $r$, $I_u$ the set of held-out items that user $u$ interacted with, and $\mathbb{I}[\cdot]$ the indicator~function.

\subsubsection{Revenue Objective.}
\label{revenue_obj}
Alongside the enhanced user experience, a company is incentivized to use a recommender to increase simultaneously the revenue. Thus, the revenue loss can be used in the training process to boost the~recommendations of expensive items and increase the overall revenue. The loss is similar to the relevance loss of Section~\ref{relevance_obj}, with a difference that the input of the model is multiplied by a weight vector, representing the prices of the items. Before computing the log-likelihood for a given user, the input vector for a user is multiplied with the price~vector:
\begin{equation}
\mathcal{L}(\boldsymbol{x};\theta,\phi) = \mathbb{E}_{q_\phi(\boldsymbol{z}|\boldsymbol{x})}[\log{p_\theta(\pi*\boldsymbol{x}|\boldsymbol{z})}]
\end{equation}
where $\boldsymbol{x}$ is the input vector for a user (it has a value  1  for  the  items  the  user  has interacted with), $\pi$ is the price vector containing the prices of each item, the $*$ symbol denotes element-wise multiplication between two vectors, $\theta$ and $\phi$ are model parameters, and $\textbf{z}$ is the variational parameter of the variational distribution.

\subsubsection{Recency Objective.}
\label{recency_obj_subsection}
From our practical experience, we came across a finding that users strongly prefer to interact with recently added content. Furthermore, the authors of \cite{ding2006recency} have shown that with the introduction of recency we could get improved and more precise recommender systems. Computing a recency score for items remains an open question. For a given dataset, we propose to leverage the timestamps of the items when they first became available. For an item, we scale its timestamp using a min-max normalization between the first and last interaction any user had with it, obtaining values in the range of [0-1]. However, we claim that recency is not a linear function of the time. Since we want to promote more recent items, we propose to transform the scores according to the following function:
    \begin{equation}
    \label{recency_eq}
        f(\rho)= 
    \begin{dcases}
        1, & \text{if } \rho\geq 0.8\\
        0.3^{(0.8-\rho)*\frac{10}{3}}, & \text{otherwise}
    \end{dcases}
    \end{equation}
The transformation function and its numeric constants have been optimized on an in-house dataset, but have been shown to generalize on other datasets. Based on this transformation function, we proposed the recency objective which stimulates the model to recommend recent items. 
The input of the model is multiplied by a weight vector, which represents the recency score of the items, when the loss is computed. Similarly to the other losses, before computing the log-likelihood for a given user, the input vector for a user is multiplied with the recency vector, or mathematically:
$
\mathcal{L}(\boldsymbol{x};\theta,\phi) = \mathbb{E}_{q_\phi(\boldsymbol{z}|\boldsymbol{x})}[\log{p_\theta(f(\rho)*\boldsymbol{x}|\boldsymbol{z})}]
$
,where $\boldsymbol{x}$ is the input vector for a user (it has a value  1  for  the  items  the  user  has interacted with), $\rho$ is the recency vector containing the recency score for each item scaled in the range of [0-1] using min-max normalization, $f$ is the function from eq. (\ref{recency_eq}), the $*$ symbol denotes element-wise multiplication between two vectors, $\theta$ and $\phi$ are model parameters, $\textbf{z}$ is the variational parameter of the distribution.

\subsection{Datasets}

In order to assess the effectiveness of our proposed model, we first carried out experiments on the well-known Amazon Books dataset, being a subsample from the Amazon review dataset \cite{he2016ups,harper2015movielens}. Along with users preferences for books, it contains the book prices which can be used as a second revenue objective (see Section~\ref{revenue_obj}). The recency is not available for this dataset.

We also consider the MovieLens 20M dataset \cite{harper2015movielens}. In terms of objectives, we use the relevance, revenue, and recency objectives for the MovieLens 20M dataset. For the revenue objective, we enriched it with prices from the Amazon review dataset by doing a fuzzy joining on the titles of the movies. For the recency objective, in the dataset for every given rating, there is a timestamp indicating when the rating was given by the user. We assume that a given movie became available when the first rating was given for it.

We train for the following combination of objectives: 1) Relevance + Revenue objectives, 2) Relevance + Recency objectives, 3) Revenue + Recency objectives, and 4) Relevance + Revenue + Recency objectives.


\subsection{Preprocessing}
In our experiments, we consider implicit feedback. We binarize ratings by converting ratings $\ge 3.5$ to positive interaction, and ratings $<3.5$ to negative interaction. Then, we split the data in a way that 90\% of the users with their interactions are used as training data, 5\% are used as validation data, and the remaining 5\% are used as testing data. Finally, we mask 20\% of interactions per user in the validation and testing data. The remaining 80\% of the interactions are used as input to the model, and the masked 20\% are used as ground truth, to compare the model's output with.

\subsection{Experimental Settings}
We implemented the state-of-the-art variational autoencoder Mult-VAE\textsuperscript{PR} of \cite{liang2018variational} for collaborative filtering, and augmented the training loss with the objectives described in Section~\ref{sec:obj_rec}. Our model contains an encoder and a decoder. The encoder consists of two linear layers of sizes 600 and 400, and the decoder also has two linear layers, both with a size of 600. The number of latent features, the bottleneck of the model is 200. We are also normalizing the input before we forward it through the model. As regularization, we use a dropout of 0.5 to the input. We will release the code. 

\subsection{Pareto Front Metrics}
It is not straightforward to compare the multi-objective solution from different multi-objective algorithms and optimization strategies. The solutions from the methods of multi-objective optimization are in the form of Pareto sets. An initial comparison of two and three-dimensional Pareto sets is to plot them and inspect them visually. Although visual inspection can help us to rank and compare Pareto set solutions, we seek an objective and systematic way. Therefore, in this section, we present three metrics for measuring the quality of the Pareto set which can help us measure the performance of the multi-objective algorithms quantitatively.\\
\indent \textbf{Hypervolume} \cite{zitzler2007hypervolume}: One of the ways of measuring the quality of the Pareto set is to measure the area that is dominated by it. The intuition is, the larger the area the solution can dominate, the better the solution. Using the hypervolume to compute the area dominated by a solution, this intuition can be extended to more than two dimensions \cite{zitzler1999multiobjective}. Since we are interested in increasing the recommender system metrics, we are using the origin as a reference point for computing the hypervolume. 


\textbf{Coverage} \cite{zitzler1999multiobjective}:
The coverage is a metric that indicates the fraction of points from one Pareto set that are dominated by or equal to points from another Pareto set. If a one point $p1$ is dominated by or equal to another point $p2$, than it is said that $p2$ covers $p1$. If the coverage is 1.0, that means all the points from the second Pareto set are covered by points from the first one. Reverse, if the coverage is 0.0, that means none of the points from the second Pareto set are covered by points from the first set. However, a drawback of the coverage is that it cannot tell us by how much one solution is better than the other one \cite{zitzler1999multiobjective}. If $P_{S_1}$ is the first Pareto set, $P_{S_2}$ is the second Pareto set, and with $p1 \geq p2$ we denote that solution point $p1$ covers solution point $p2$, then the coverage metric is defined as:
$
\mathcal{C}(P_{S_1}, P_{S_2}) = \frac{|\{p2 \in P_S2; \exists p1 \in P_{S_1}: p1 \geq p2\}|}{|P_{S_2}|}
$

It is important to note that the coverage metric is not symmetric, and both $\mathcal{C}(P_{S_1}, P_{S_2})$ and $\mathcal{C}(P_{S_2}, P_{S_1})$ have to be examined when evaluating Pareto sets. In our experiments, we report both variants as we apply a pairwise comparison.

\textbf{Spacing} \cite{okabe2003critical}:
The spacing is a distance-based metric that measures the spread of a given solution. The bigger the spacing metric is, the more diverse and the more spread are the solutions in the Pareto set. If having the best solutions in a Pareto set is important, the diversity of the solutions captures the range of choices available to the decision-makers. This is a concrete business advantage. If $P_S$ is the Pareto set, $d_i$ is the distance to the closest neighbour of the $i$-th point in the Pareto set, and $\bar{d}$ is the average of $d_i$, then the spacing is computed as:
$
\mathcal{SP}(P_S) = \sqrt{\frac{1}{|P_S|-1} \sum_{i=1}^{|S|} {(d_i - \bar{d})^2}}
$

\section{Results}

\subsection{Two Objectives}


\begin{figure*}[!t]
    \centering
    \subfloat[MovieLens 20M dataset.]{{\includegraphics[width=0.30\textwidth,height=3cm]{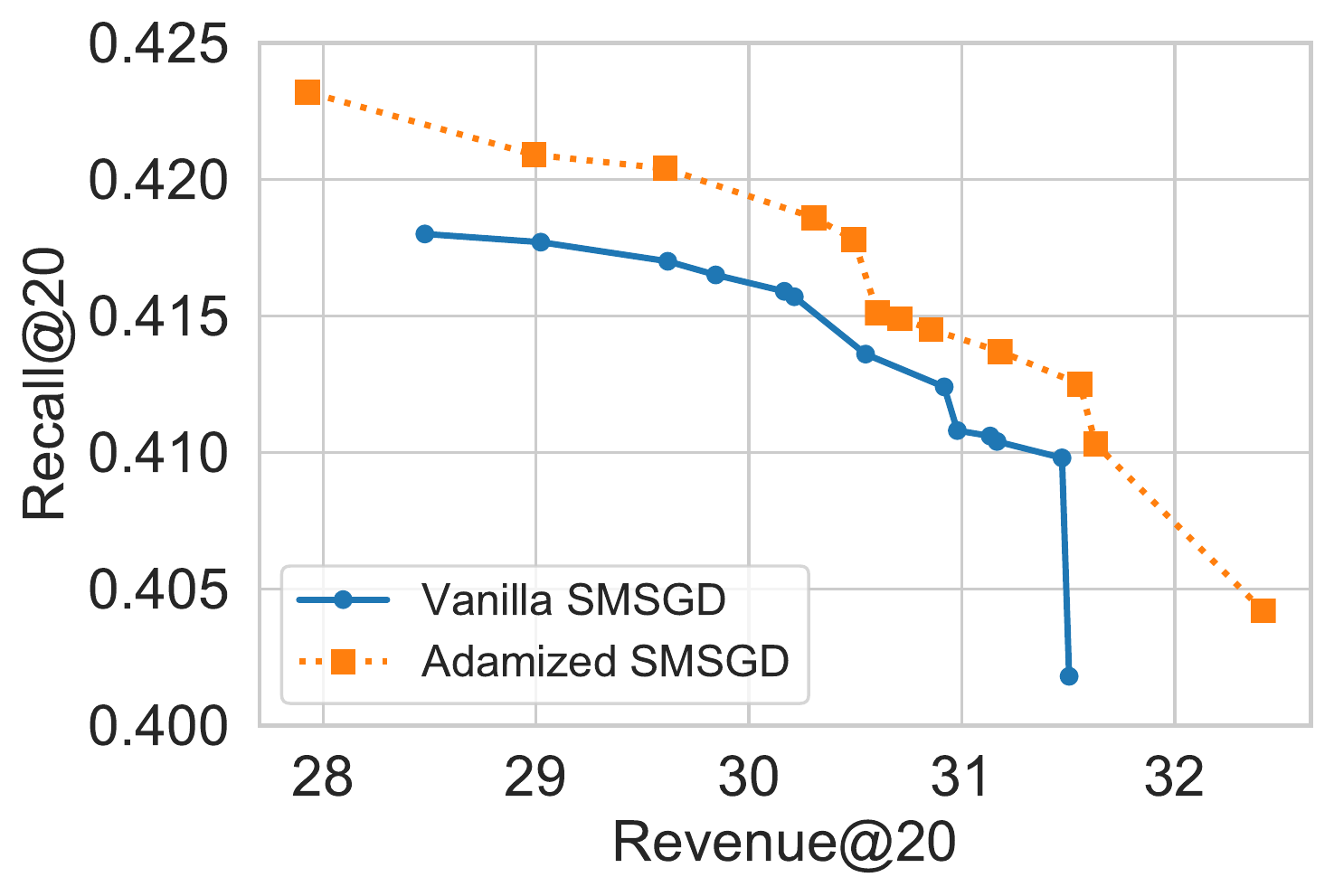} }}
    \subfloat[Amazon dataset.]{{\includegraphics[width=0.30\textwidth,height=3cm]{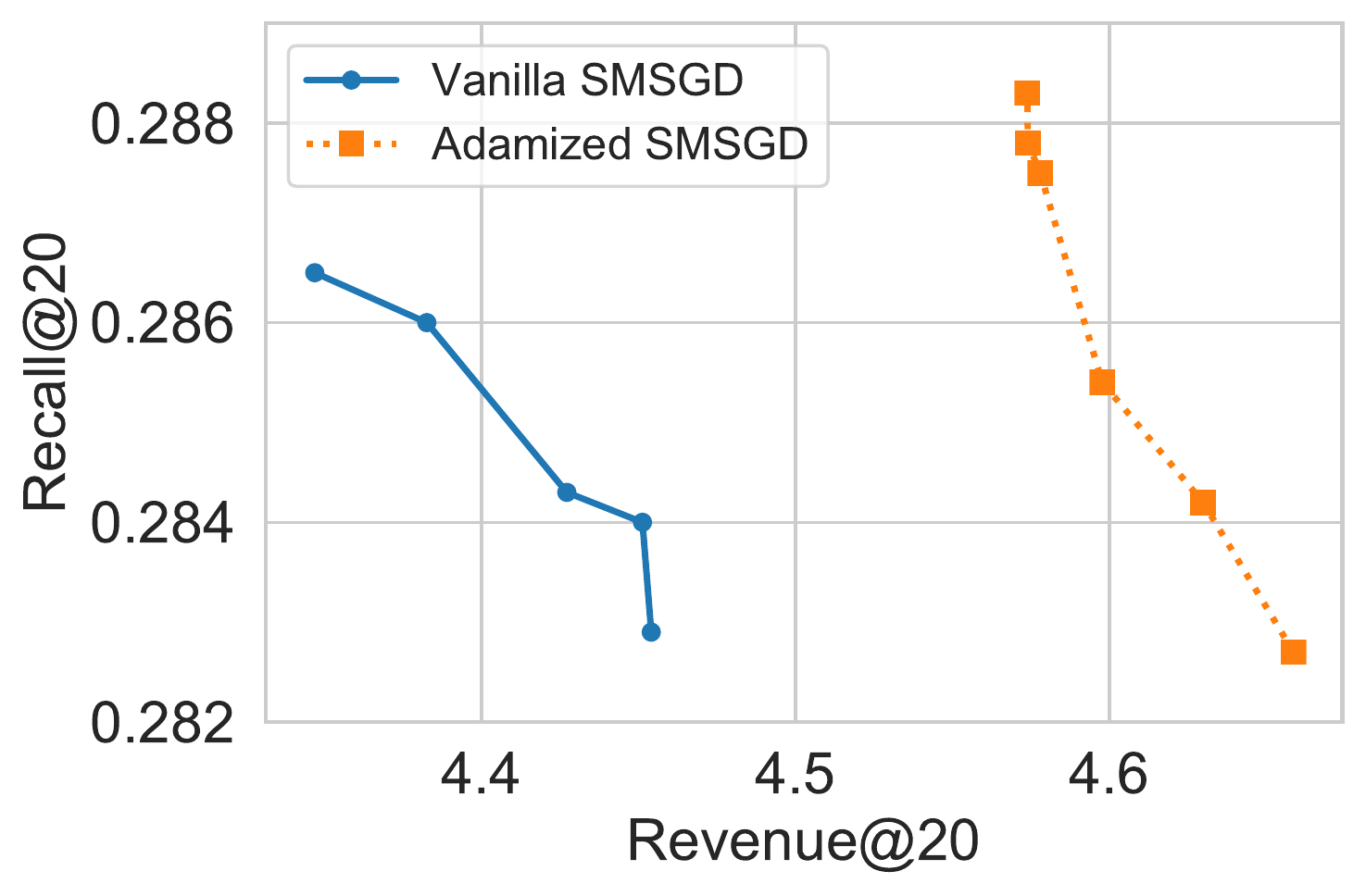} }}%
    \caption{Visualization of Pareto fronts for two objectives.}%
    \label{fig:PF_adamize}%
    \Description{The results in a graph format. We have shown the Pareto fronts for two objectives on the MovieLens 20M and Amazon datasets.}
\end{figure*}
Figure~\ref{fig:PF_adamize} shows the Pareto front of the baseline and our method. From both visualizations we can clearly observe that \textit{Adamizing} the gradients substantially improves the performance over the SMSGD algorithm. The Pareto front obtained with our method clearly dominates the vanilla SMSGD algorithm. On MovieLens, the Pareto fronts are more spread than in the Amazon dataset case.\\
\indent To further inspect and quantify the results, we also present the metrics for measuring the quality of the Pareto set in Table~\ref{adamize_metrics_table}. Our proposed algorithm outperforms the baseline substantially in terms of coverage (as can be seen on the visualization) also in terms of hypervolume, following the visualiation. However, we observe that the spacing of our method nearly doubles in the MovieLens dataset, but perform similarly on the Amazon Book dataset.


\subsection{Three Objectives}

\begin{figure*}[!t]
    \centering
    \subfloat[Revenue vs Recall.]{{\includegraphics[width=0.30\textwidth,height=3cm]{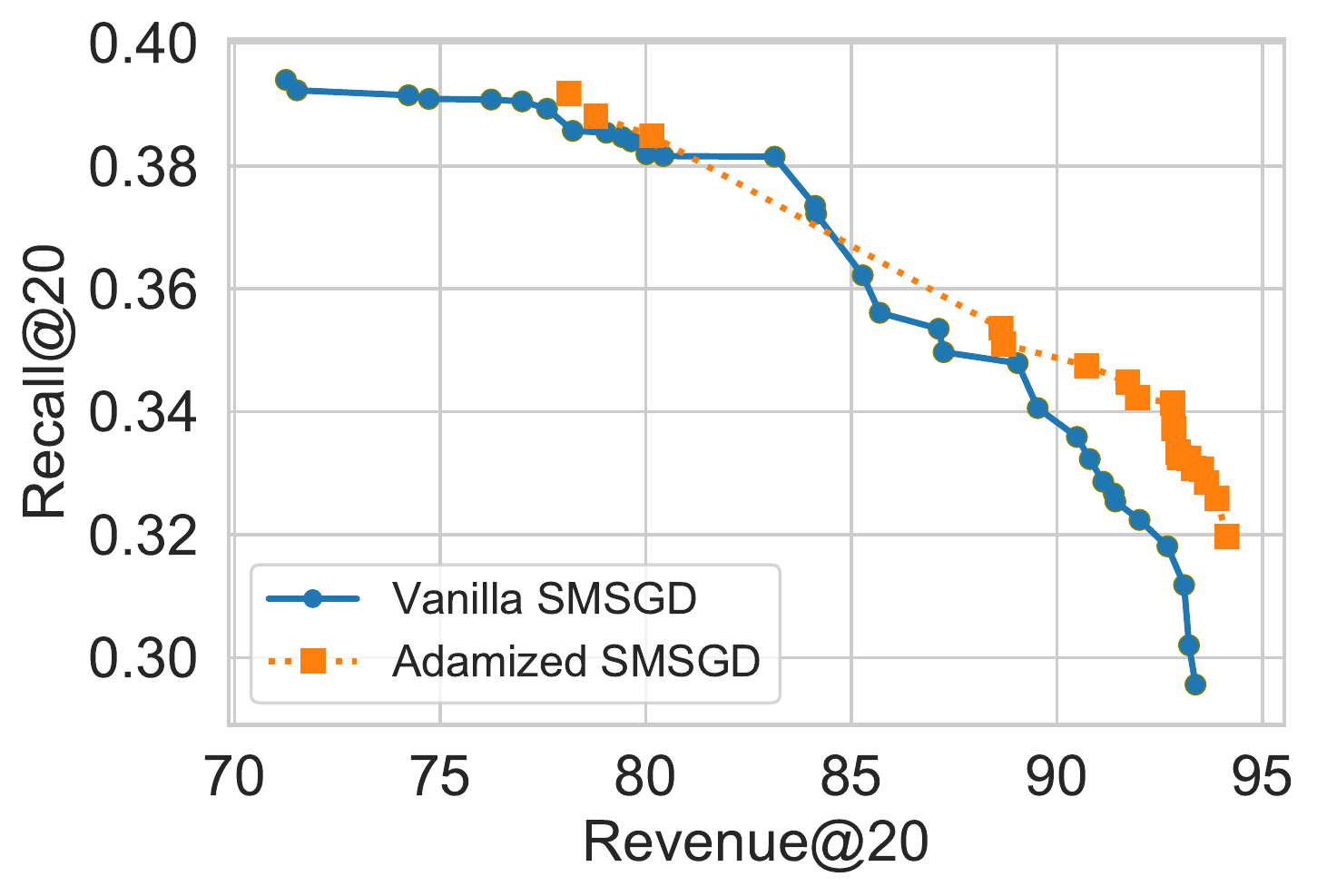} }}%
    \subfloat[Recency vs Recall.]{{\includegraphics[width=0.30\textwidth,height=3cm]{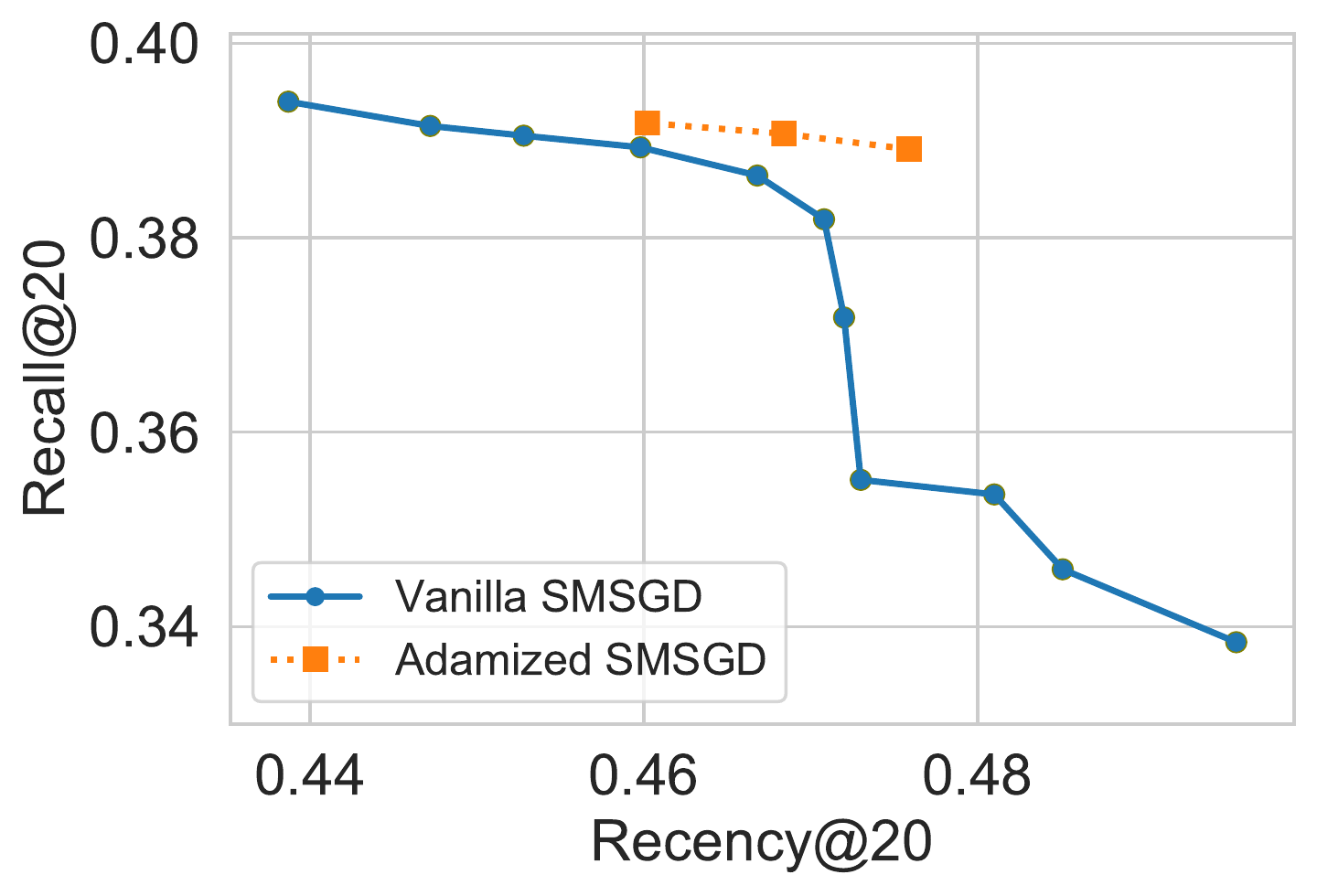} }}%
    \subfloat[Recency vs Revenue.]{{\includegraphics[width=0.30\textwidth,height=3cm]{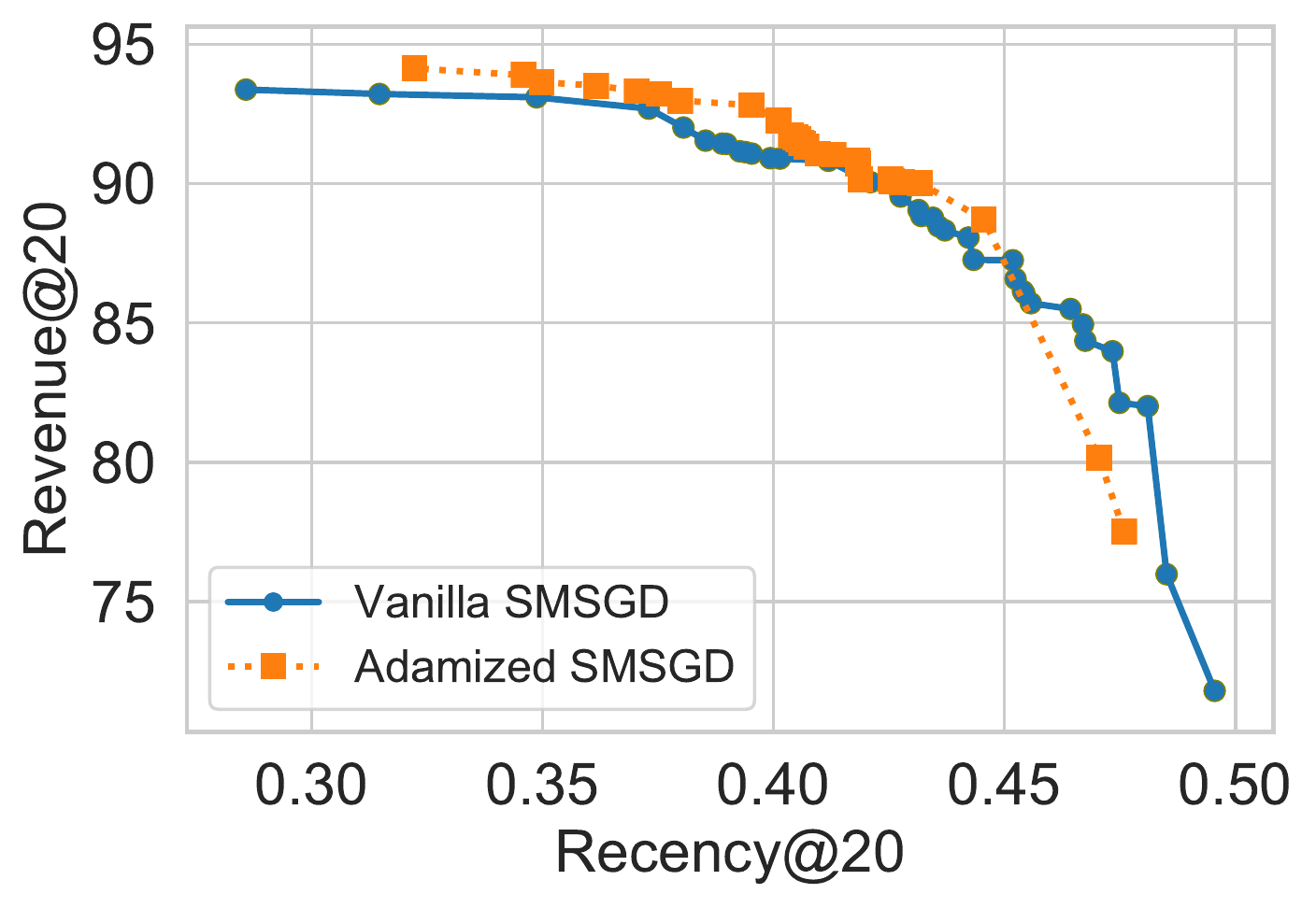} }}%
    \caption{Pareto fronts for MovieLens on three objectives.}%
    \label{fig:PF_adamize_three_obj}%
    \Description{The results in a graph format. We have shown the Pareto fronts for three objectives on the MovieLens and Amazon datasets, but projected on two dimensions by ignoring one objective on every plot. More precisely, on a) we have plotted the revenue vs recall, on b) we have plotted recency vs recall, and on c) we ave plotted recency vs revenue.}
\end{figure*}

For better visualization, we project the three-dimensional Pareto fronts on two objectives. Results are available in Figure~\ref{fig:PF_adamize_three_obj}. Still, from the plots we observe an improvement in all the three combination of objectives.

The Table~\ref{adamize_metrics_table_three_obj} quantifies the improvement of our proposed method compared to the vanilla SMSGDA. We can see that the \textit{Adamize} trick on our method dominates approximately half the solutions found by the vanilla SMSGDA, while being slightly more spread over the space. In terms of hypervolume, the vanilla SMSGDA performs slightly better. However, the difference is less significant than the two objectives case because of the curse of dimensionality.

Supported by the improvements on two different datasets, and using up to three different objectives, we can say that the \textit{Adamize} trick leads on average to better solutions.

\begin{minipage}{0.46\textwidth}
    \begin{table}[H]
    \small
    \centering
    \captionsetup{justification=centering}
    \caption{Pareto front metrics for datasets with two objectives.}
    \begin{tabular}{@{}clccc@{}}
    \textbf{Dataset} & \textbf{Method} & \textbf{Hypervolume} & \textbf{Coverage} & \textbf{Spacing} \\
    \midrule
    \multirow{2}{*}{Movies} & Vanilla & 13.16                & 0.0               & 0.19\\
    & Adamized & \textbf{13.68} & \textbf{1.0}      & \textbf{0.34}    \\
    \hline
    \multirow{2}{*}{Books} & Vanilla  & 1.28                 & 0.0               & \textbf{0.016}   \\
    & Adamized & \textbf{1.34}        & \textbf{1.0}      & 0.014           
    \end{tabular}
    \label{adamize_metrics_table}
    \end{table}
\end{minipage}
\hfill
\begin{minipage}{0.46\textwidth}
    \begin{table}[H]
    \small
    \centering
    \captionsetup{justification=centering}
    \caption{Pareto front metrics for MovieLens on three~objectives.}
    \begin{tabular}{@{}lccc@{}}
    \textbf{Method} & \textbf{Hypervolume} & \textbf{Coverage} & \textbf{Spacing}\\
    \midrule
    Vanilla  & \textbf{17.54}                & 0               & 0.15             \\
    Adamized & 17.02       & \textbf{0.49}      & \textbf{0.24}    \\
    \end{tabular}
    \label{adamize_metrics_table_three_obj}
    \end{table}
\end{minipage}

\section{Conclusion}
In this paper we introduced a novel method for multi-gradient descent that leverages a momentum-based optimizer. We applied the method on a problem with a growing importance - Multi-objective Recommender Systems. We benchmarked the novel optimization method against the state-of-the-art multi-gradient descent method and reported the results on three different metrics based on the resulting Pareto front: \textit{hypervolume}, \textit{coverage}, and \textit{spacing}. The results show that the new Pareto fronts are substantially better from all three perspectives. We complemented the analysis with a visualization of the Pareto fronts that further emphasizes the gains obtained.\\
\indent To the best of our knowledge, we are the first to use a momentum-based optimizer for each objective in a multi-objective setup. We hope that this will inspire research practitioners to test and produce other ideas in the direction of using momentum-based optimizers per objective in a multi-objective setup. Improving the gradient-based optimization could benefit all the multi-objective optimization problems, in all applicable fields.

\bibliographystyle{ACM-Reference-Format}

\end{document}